\documentclass[10pt,twocolumn,letterpaper]{article}

\usepackage{iccv}
\usepackage{times}
\usepackage{epsfig}
\usepackage{graphicx}
\usepackage{amsmath}
\usepackage{amssymb}

\usepackage{xcolor}         % colors
\usepackage{xspace}
\usepackage{amsmath,amssymb,amsfonts,dsfont,pifont,bm,bbm,mathrsfs,mathtools,nicefrac}
\usepackage{wrapfig}
\usepackage{booktabs,multirow,adjustbox,diagbox,threeparttable}
\usepackage{amsfonts}
\usepackage{bbm}

\usepackage[pagebackref=true,breaklinks=true,letterpaper=true,colorlinks,bookmarks=false]{hyperref}

\usepackage[capitalize]{cleveref}
\crefname{section}{Sec.}{Secs.}
\Crefname{section}{Section}{Sections}
\Crefname{table}{Table}{Tables}
\crefname{table}{Tab.}{Tabs.}

\iccvfinalcopy % *** Uncomment this line for the final submission

\def\iccvPaperID{3116} % *** Enter the ICCV Paper ID here

\ificcvfinal\fi

\begin{document}

%%%%%%%%% TITLE
\title{Informative Data Mining for One-shot Cross-Domain Semantic Segmentation}

%\author{First Author\\
%Institution1\\
%Institution1 address\\
%{\tt\small firstauthor@i1.org}
% For a paper whose authors are all at the same institution,
% omit the following lines up until the closing ``}''.
% Additional authors and addresses can be added with ``\and'',
% just like the second author.
% To save space, use either the email address or home page, not both
%\and
%Second Author\\
%Institution2\\
%First line of institution2 address\\
%{\tt\small secondauthor@i2.org}
%}

%\author{{Yuxi Wang$^{1,2}$, \quad Jian Liang$^{2}$, \quad Jun Xiao, \quad Shuqi Mei, \quad Yuran Yang, \quad Zhaoxiang Zhang} \\
%{
%$^{1}$ Centre for Artificial Intelligence and Robotics, HKISI\_CAS \\
%$^{2}$ Institute of Automation, Chinese Academy of Sciences \\
%$^{3}$ University of Chinese Academy of Sciences \\
%$^{4}$ State Key Laboratory of Multimodal Artificial Intelligence Systems 
%$^{5}$ Tencent \\
%}
%{\tt\small secondauthor@i2.org}
%}

\author{
    Yuxi~Wang$^{1,2}$ \hspace{0.6mm}
    Jian~Liang$^{2,3,4}$ \hspace{0.6mm}
    Jun~Xiao$^{3}$ \hspace{0.6mm}
    Shuqi~Mei$^{5}$ \hspace{0.6mm}
    Yuran~Yang$^5$ \hspace{0.6mm}
    Zhaoxiang~Zhang$^{1,2,3,4,}$\thanks{ Zhaoxiang Zhang is the corresponding author.} \\%[8pt]
    $^1$Centre for Artificial Intelligence and Robotics, HKISI-CAS \\ %Hong Kong Institute of Science \& Innovation, Chinese Academy of Sciences \\ %HKISI\_CAS \\%[3pt] \\
    $^2$Institute of Automation, Chinese Academy of Sciences\,
    $^3$University of Chinese Academy of Sciences \\
    $^4$State Key Laboratory of Multimodal Artificial Intelligence Systems 
    $^5$Tencent \\
    \small{\texttt{\{yuxiwang93, liangjian92\}@gmail.com}} \quad
    \small{\texttt{zhaoxiang.zhang@ia.ac.cn}}
}

\maketitle
% Remove page # from the first page of camera-ready.
%\ificcvfinal\thispagestyle{empty}\fi

%\footnote{$^*$ Corresponding Author}

\begin{abstract}
\label{Sec:abstract}

    Contemporary domain adaptation offers a practical solution for achieving cross-domain transfer of semantic segmentation between labelled source data and unlabeled target data. These solutions have gained significant popularity; however, they require the model to be retrained when the test environment changes. This can result in unbearable costs in certain applications due to the time-consuming training process and concerns regarding data privacy. One-shot domain adaptation methods attempt to overcome these challenges by transferring the pre-trained source model to the target domain using only one target data. Despite this, the referring style transfer module still faces issues with computation cost and over-fitting problems.
    To address this problem, we propose a novel framework called Informative Data Mining (IDM) that enables efficient one-shot domain adaptation for semantic segmentation. Specifically, IDM provides an uncertainty-based selection criterion to identify the most informative samples, which facilitates quick adaptation and reduces redundant training. We then perform a model adaptation method using these selected samples, which includes patch-wise mixing and prototype-based information maximization to update the model. This approach effectively enhances adaptation and mitigates the overfitting problem.
    In general, we provide empirical evidence of the effectiveness and efficiency of IDM. Our approach outperforms existing methods and achieves a new state-of-the-art one-shot performance of 56.7\%/55.4\% on the GTA5/SYNTHIA to Cityscapes adaptation tasks, respectively. The code will be released at \url{https://github.com/yxiwang/IDM}. 

\end{abstract}

\section{Introduction}\label{sec:intro}

    Semantic segmentation is a fundamental computer vision task that has made remarkable progress with the help of vast amounts of pixel-level annotated training data. However, collecting such large-scale datasets requires tremendous labeling efforts in terms of time and cost. For instance, annotating a single image in the Cityscapes dataset can take about 90 minutes and cost 1.5 dollars \cite{Cityscapes}. To reduce this burden of labeling, previous cross-domain semantic segmentation approaches \cite{FCN,AdaptSeg,CLAN,ADVENT,FDA,BDL,CAG,BiMAL} have been developed. These methods transfer the knowledge from label-rich synthetic data (source) \cite{GTA5,SYNTHIA} to the unlabeled real-world data (target) \cite{Cityscapes}, referred to as domain adaptive semantic segmentation (DASS).

    %%%%%%%%%%%%%%%%%%%%%%%%%%%%%
    \begin{figure}[!tbp]
        \centering
        \includegraphics[width=1.0\linewidth,trim=145 130 250 170,clip]{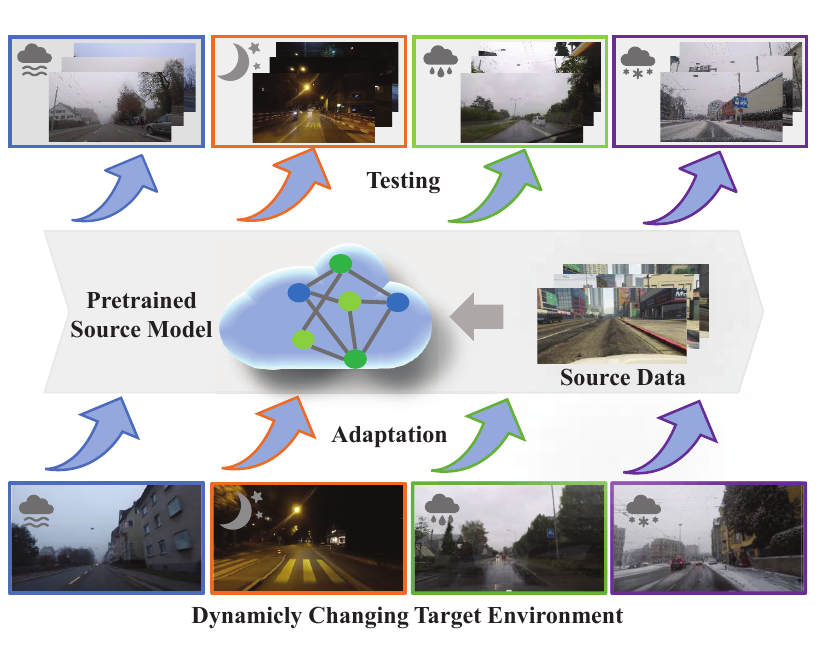}
        \caption{
        We consider the one-shot domain adaptation scenario, where only one single target image is used to fit the trained source model. It is realistic for the adaptation model to tackle dynamically changing target environments, such as suddenly occurrent weather (\textit{e.g.} foggy, rainy, night, snow). Our method aims to achieve quick adaptation for one-shot domain adaptation.}
        \label{Fig:minmax}
    \end{figure}
    %%%%%%%%%%%%%%%%%%%%%%%%%%%%%

    Despite significant efforts in developing DASS methods, most of them still suffer from the following limitations. First, existing approaches train an adaptation model from a specific source domain to a specific target domain. Therefore, they require retraining the model every time when the test environment changes, which is inflexible and inefficient in handling a dynamic domain shift scenario. 
    Second, these methods need access to the entire target dataset to achieve model adaptation, which is impractical in some realistic scenarios due to privacy or storage concerns.
    As depicted in Figure \ref{Fig:minmax}, an autonomous driving model often faces sudden weather or illumination changes, resulting in a scarcity of images at the beginning of environmental shifts. Therefore, it is crucial for the model to rapidly adapt to the new conditions with a limited amount of target data. Although previous works \cite{ASM,SM-PPM} have attempted to address the one-shot domain adaptation problem, the introduced style transfer module usually requires additional style images by an extra optimizing model. This makes it inefficient to adapt quickly to different scenarios.
    
    In contrast to pioneering works, our method provides a new direction for OSDA by exploring the abundant information hidden in the source domain due to the rare target data accessible. 
    To achieve this goal, two key ideas exist to address this problem. First, we select the most informative samples for training, which can reflect the target distribution and reduce the redundant back-propagation process. Second, we diversify the target distribution to alleviate the over-fitting problem caused by the limited availability of one-shot target data.
    To this end, we propose the Informative Data Mining (\textit{\textbf{IDM}}) approach, which includes a sample selection strategy and an efficient model adaptation technique. Specifically, we first introduce an uncertainty-based selection criterion to identify the most informative training samples from the source data. These samples can reflect the target distribution and contribute significantly to model adaptation. Therefore, the sample selection criteria aim to filter the most informative images with 1) higher prediction uncertainty values and 2) higher diversity. Unlike existing works choosing low-uncertainty target samples to refine the target model, our method filters target-style-like source images for training.
    On the other hand, we devise a model adaptation method seeking to alleviate the over-fitting problem by diversifying the distribution of the target domain. Concretely, we first diversify the semantic content of target data by a patch-wise mixing method and then use prototype-based information maximization to ensure the output of diversity for the selected source data, which can significantly alleviate the over-fitting problem. 
    With the proposed IDM method, the model can be efficiently adapted to the target data without over-fitting.

    Our contributions can be summarized as follows:
    %\begin{enumerate}
        1) We propose a new efficient one-shot adaptation framework for cross-domain semantic segmentation, which aims at quickly adapting the trained source model to the target data with only hundreds of training iterations.
        2) We propose a novel sample selection scheme to filter out the most informative training samples for reducing redundant optimization and devise an uncertainty minimization training technique for model adaptation.
        3) We show the efficacy and efficiency of our method by achieving a new state-of-the-art performance on one-shot domain adaptive semantic segmentation, with 56.7\% mIoU on GTA5 to Cityscapes and 55.4\% mIoU on SYNTHIA to Cityscapes. 
    %\end{enumerate}

\section{Related work}\label{sec:related}

\textbf{Unsupervised Domain Adaptive Semantic Segmentation}
    aims to transfer the pixel-level annotations from the source domain to the target domain. Existing approaches can be roughly categorized into two groups: adversarial learning based method and self-training based method. For adversarial learning, numerous works focus on reducing the distribution misalignment in the image level \cite{CrDoCo,TGCF-DA,CyCADA,sankaranarayanan2018learning,ACE,FDA,TextureDA}, feature level \cite{chang2019all,hong2018conditional,MaxSquare,fan2022toward,hu2021adversarial}, and output level \cite{AdaptSeg,CLAN,ADVENT,PatchDA,zhao2017survey}. For the self-training based method, the essential idea is to generate reliable pseudo labels. Typical approaches usually consist of two steps: 1) generate pseudo labels based on the source model \cite{CBST,CRST,UncerDA,U2PL,ATP,wang2022self,tian2023self} or the learnt domain-invariant model \cite{Rectifying,PyCDA,LSE,ProDA}, 2) refine the target model supervised by the generated pseudo labels. PLCA \cite{PLCA, T2S-DA} introduces a new paradigm for domain adaptive semantic segmentation via building pixel-level cycle association between source and target pixel pairs. 
    Although these methods have achieved promising results, they usually access amounts of source and target data. In this paper, we address a challenging and practical target data-scarce setting where only a one-shot unlabeled target image is available during adaptation.

\begin{figure*}[!t]
    \centering
    \includegraphics[width=0.98\textwidth,trim=90 120 60 220,clip]{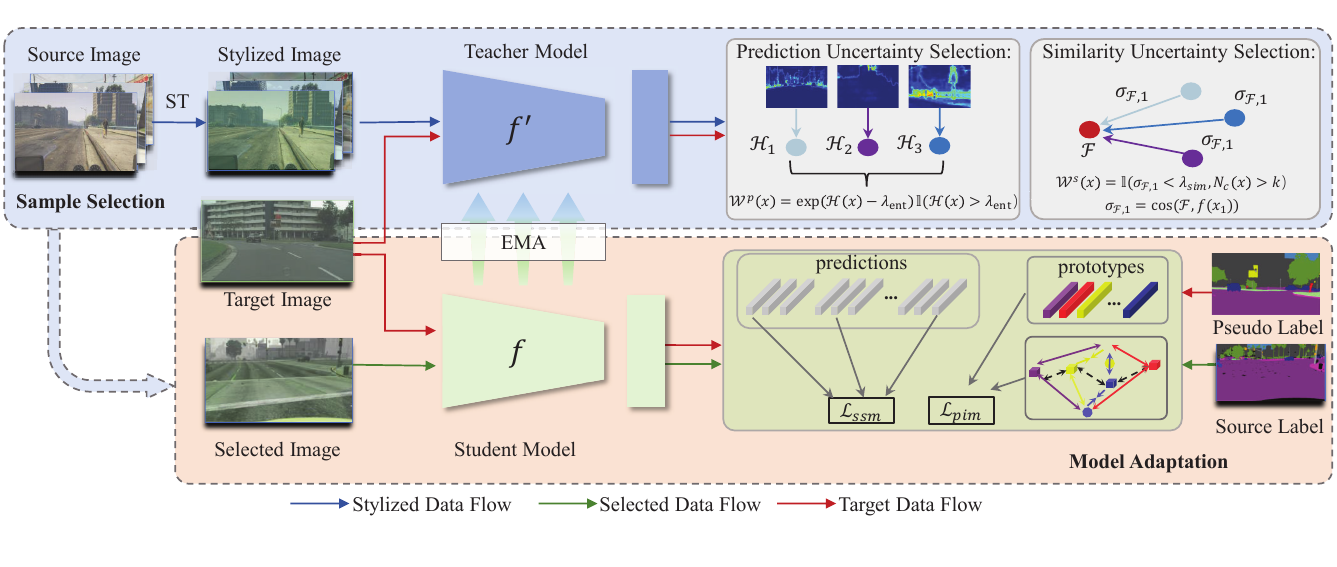}
    \caption{
        \textbf{An overview of our proposed IDM.} To achieve one-shot domain adaptation, our method consists of two strategies. 1) Sample selection aims to identify the most informative images to optimize the model. It first generates target-like stylized images by the style transfer (ST) technique. Then, the stylized images are fed into the teacher model to select the trained data by the proposed prediction and similarity uncertainty selection techniques; 2) The model adaptation aims to update the segmentation model on the selected and given target data. In this process, we update the model by the proposed $\mathcal{L}_{pim}$ and $\mathcal{L}_{ssm}$ to alleviate over-fitting problem. The teacher model is initialized by the pretrained source model, and it is updated in the EMA manner. 
    }
    \label{fig:pipeline}
    %\vspace{-10pt}
\end{figure*}

    \noindent\textbf{One-shot Domain Adaptation (OSDA)} aims to overcome the need for larger training sets and improve the capability of transferring the trained source model to a new target domain with having access to the source data and only one target data. 
    Recently, OSDA has achieved significant progress in dealing with face generalization \cite{yang2020one}, object detection \cite{wan2020one}, and semantic segmentation \cite{ASM,SM-PPM}. For semantic segmentation, ASM \cite{ASM} proposes an adversarial style mining algorithm by mutually optimizing the style-transfer module and the segmentation network via an adversarial regime. \cite{SM-PPM} integrates a style-mixing technique into the segmentor to stylize the source images without introducing any learned parameters. S4T \cite{S4T,liang2023comprehensive} is another representative method to design a regularized self-learning signal at test-time, which proposes a selective self-training scheme for semantic segmentation by regularizing pseudo labels with aligned predictive view generation. On the contrary, we aim to mine the most informative images to achieve quick adaptation and reduce the over-fitting problem.  

\noindent\textbf{Domain Generalized Semantic Segmentation (DG)}
    has attracted considerable attention in recent years, which aims to learn a generalized model on the source domain and performs well on a novel domain. To improve performance on novel domains, most existing studies focus on whitening \cite{ISW}, normalizing \cite{IBN}, and diversifying \cite{peng2021global,FDSR,DRPC} styles to avoid over-fitting to the style of the source domain. Domain generalized semantic segmentation is close to our work, which assumes target data is inaccessible during training. The difference is that no target data is available in DG, while one target image is accessible in our work. 

\section{Method}
\label{sec:method}

    In this section, we formally introduce the proposed Informative Data Mining (\textbf{IDM}) method that aims to improve the efficiency of one-shot domain adaptive semantic segmentation. In this setting, we only access one unlabeled target image $\mathit{x}_t \in \mathcal{X}_t$, and $n_s$ source data $\{x_s^i, y_s^i\}_{i=1}^{n_s} = \{\mathcal{X}_s, \mathcal{Y}_s\}$. As shown in Figure \ref{fig:pipeline}, 
    the proposed \textbf{IDM} consists of two strategies, 1) sample selection step and model adaptation step. The former attempts to identify the most informative images for training, and the latter aims to achieve quick adaptation without over-fitting. We assume the informative data should have two properties: contributing more to adaptation and reducing redundant optimization. Therefore, we propose prediction and similarity uncertainty selection techniques to filter the most informative training samples. After that, model adaptation seeks to alleviate the over-fitting problem by diversifying the distribution of the target domain, following the direction of the given target data.

    \subsection{Sample Selection}
    \label{sec:SEM}
    
    To achieve one-shot domain adaptation quickly, detecting the most informative samples that can reflect the target distribution for backward propagation is essential. Therefore, we propose a sample selection strategy considering the following two criteria: 1) samples should be \textit{target-style closer}, and 2) the distributions of involved samples should be \textit{diverse}, including various distributions and categories. 
    
    \noindent \textbf{Prediction Uncertainty Selection.}  
    Different from previous works selecting target samples with lower uncertainty to refine the model \cite{niu2022efficient}, our method aims to filter the target-like source images for adaptation. Although images with lower uncertainty reveal reliable predictions,  they often refer to source-like images rather than target-like ones. Therefore, we filter uncertain samples and assign them higher training weights as they contribute more to adaptation.
    The motivation lies in the following two aspects. (a) \textit{The initialized model is trained on the source data.} (b) \textit{Fine-tuning on underperforming samples (higher uncertainty) is more valuable}. Due to (a), target-like samples generally perform higher prediction uncertainty than source-like ones, which can alleviate the confusion of higher uncertainty only belonging to ``hard samples". As (b), we should select higher uncertainty samples that have ground truth labels (source labels). 
    Besides, target-like samples have high-uncertainty predictions on the source model, but not all images with high-entropy predictions are similar to the target domain. To alleviate this confusion, we apply the style transfer technique \cite{AdaIN, FDA} to generate target-style images by treating the given target image as an ``anchor style". Then the prediction uncertainty selection is performed as:
    \begin{equation}
        \label{Eq:entropy_es}
        \footnotesize
        \mathcal{W}^{p}(\hat{\mathrm{x}}_s) = \exp (\mathcal{H}(\hat{\mathrm{x}}_s) - \lambda_{ent}) \cdot \mathbb{I}(\mathcal{H}(\hat{\mathrm{x}}_s) > \lambda_{ent}), 
    \end{equation}
    where $\mathbb{I}(\cdot)$ is an indicator function, $\lambda_{ent}$ denotes a pre-defined threshold, and $\mathcal{H}(\hat{\mathrm{x}}_s)$ is the mean entropy of stylized source image $\hat{\mathrm{x}}_s$. For $\mathrm{\hat{x}}_s$ generation, we transfer the style of the target image $\mathrm{x}_t$ to the source data following \cite{AdaIN,DSU}: 
    \begin{equation}
        \label{eq:mix}
        \footnotesize
        \hat{\mathrm{x}}_{s} = \beta(\mathrm{x}_t) (\frac{\mathrm{x}_s - \mu(\mathrm{x}_s)}{\sigma(\mathrm{x}_s)}) + \gamma(\mathrm{x}_t).
    \end{equation}
    $\mu(\mathrm{x}_s)$ and $\sigma(\mathrm{x}_s)$ are the mean and standard deviation of source images. $\beta(\mathrm{x}_t)$ and $\gamma(\mathrm{x}_t)$ are the reconstructed target statistic of mean and standard deviation, formulating as:   
    \begin{equation}
        \label{eq:mu_target}
        \footnotesize
        \begin{aligned}
        & \gamma(\mathrm{x}_t) = \mu(\mathrm{x}_t) + \delta_{\mu} \| \mu(\mathrm{x}_t)-\mu(\mathrm{x}_s)) \|, \\
        & \beta(\mathrm{x}_t) = \sigma(\mathrm{x}_t) + \delta_{\sigma} \| (\sigma(\mathrm{x}_t)-\sigma(\mathrm{x}_s)) \|,
        \end{aligned}
    \end{equation}
    where $\delta_{\mu}$ and $\delta_{\sigma}$ control the weights of statistic offset and they are randomly sampled from Gaussian $\mathcal{G}(0,1)$.

    \noindent \textbf{Similarity Uncertainty Selection.} Although Eq. (\ref{Eq:entropy_es}) identifies the most informative samples with higher prediction uncertainty, the restriction may still be limited. For example, selected images may only have some frequent classes, harming rare classes' performance. Moreover, the two selected samples have similar representations, with both performing a higher prediction entropy than $\lambda_{ent}$. Therefore, it is redundant for optimization as they produce an equal contribution to back-propagation.           
    
    To address this problem, we exploit the samples with diverse representations and categories in this section. Specifically, we first ensure the involved images contain different categories and then guarantee the output of selected images is not similar. Since calculating the similarity between the current image and all filtered samples is time-consuming and computationally expensive, inspired by \cite{niu2022efficient}, we conduct a memory bank to store the average outputs of selected samples, denoted as $\mathcal{F}$. 
    Then the similarity uncertainty selection is formulated as follows:
    \begin{equation}
        \label{Eq:diver_es}
        \footnotesize
        \mathcal{W}^{s}(\hat{\mathrm{x}}_s) =  \mathbb{I}(\mathit{cos}(f(\mathrm{\hat{x}_s}), \mathcal{F}) < \lambda_{sim}, N_c(\mathrm{\hat{x}_s}) > k),
    \end{equation}
    where $\mathit{cos}(\cdot, \cdot)$ is the cosine similarity operator, $\mathit{f}$ is the segmentation network, $\lambda_{sim}, k$ are pre-defined thresholds, and $N_c(\mathrm{\hat{x}}_{s})$ indicates the number of classes that $\mathrm{\hat{x}}_s$ contains.  
    
    Based on \textit{prediction uncertainty selection} $\mathcal{W}^{p}(\mathrm{\hat{x}}_s)$ and \textit{similarity uncertainty selection} $\mathcal{W}^{s}(\mathrm{\hat{x}}_s)$, we can obtain an overall sample-selection weight as:
    \begin{equation}
        \footnotesize
        \mathcal{W}(\mathrm{\hat{x}}_s) = \mathcal{W}^{p}(\mathrm{\hat{x}}_s) \cdot \mathcal{W}^{s}(\mathrm{\hat{x}}_s).
    \end{equation}

    The selected samples are used to optimize the segmentation model by minimizing the following objective:
    \begin{equation}
        \label{equ:sem}
        \footnotesize
        \mathcal{L}_{ssm}(\mathrm{\hat{x}}_s) = - \mathcal{W}(\mathrm{\hat{x}}_s) \sum_{i=1}^{N} \sum_{j=1}^{H\times W}\mathrm{y}_s^{(i,j)}\log [p(\mathrm{\hat{x}}_s^{(i,j)})],
    \end{equation}
    where $N$ is the number of selected images to fine-tune the adaptation model. $H$ and $W$ denote the height and width of the image. $p(\mathrm{\hat{x}}_s^{(i,j)})$ indicates the probabilistic output of the $j$-th pixel for the $i$-th image $\mathrm{\hat{x}}_s$. $\{\mathrm{\hat{x}}_s, \mathrm{y}_s\}$ is the training pair with a stylized source image and ground truth label. 
    
    Benefiting from the sample selection, we can select the most informative samples for training and reduce redundant optimization, helping to achieve quick adaptation for OSDA. Note that the sample-selected process does not involve any gradient back-propagation. Therefore, it is impressive to save computing resources.

    %%%%%%%%%%%%%%%%%%%%%%%%%%%%%
    \begin{figure}[!tbp]
        \centering
        \includegraphics[width=1.0\linewidth,trim=70 225 350 155,clip]{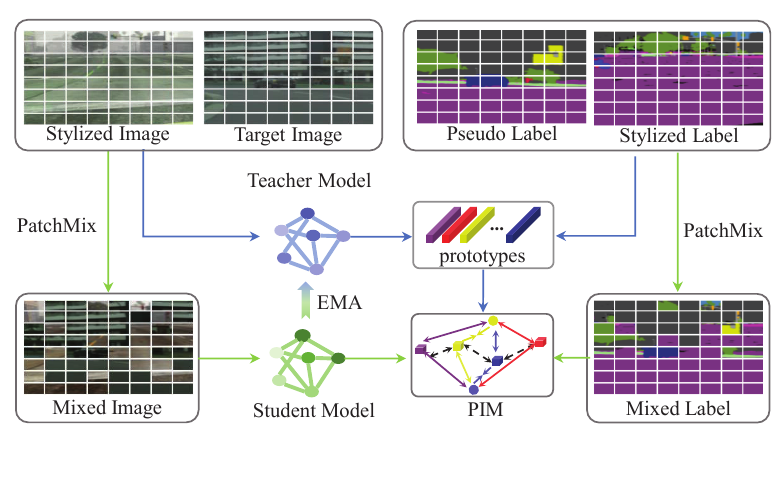}
        \caption{
        Illustration of the model adaptation process. PIM indicates the module of prototype-based information maximization.}
        \label{Fig:patchmax}
    \end{figure}
    %%%%%%%%%%%%%%%%%%%%%%%%%%%%%
    
    \subsection{Model Adaptation}
    \label{sec:DDM}
    
    Considering one-shot target images available, it only provides biased style and content information for the target domain. Despite previous works \cite{ASM, SM-PPM} using style transfer to estimate the target distribution, the generated images still have a bias to the real dataset. Besides, the sole target cannot correctly reflect the target distribution, and it is easy to overfit existing categories. To alleviate this problem, we propose a novel data augmentation technique to diversify the target data. Specifically, we devise a patch-wise mixing between the selected stylized image and the given target data, which seeks to explore the content and style diversity for the training images. Then we use these mixed images to conduct a prototype-based information maximization to ensure the diversity of predictions.
    
    \noindent\textbf{Patch-wise Mixing.} Class mixing \cite{ClassMix,DACS} is an excellent technique to improve adaptation performance. However, in the one-shot scenario, it is inapplicable because most categories are missing in the target data. Therefore, we introduce a patch-wise mixing method by splitting the target and stylized source images into $P$ patches. We obtain a new mixed image by randomly replacing the source patch with the target. The corresponding labels are obtained in the same way:
    \begin{equation}
        \label{Eq:PatchMix}
        \footnotesize
        \mathrm{\widetilde{x}}_t = \mathrm{\textbf{PatchMix}}(\mathrm{\hat{x}}_s, \mathrm{x}_t), \mathrm{\widetilde{y}}_t = \mathrm{\textbf{PatchMix}}(\mathrm{{y}}_s, \mathrm{y}_t'),
    \end{equation}
    where $\mathrm{\textbf{PatchMix}}(\cdot, \cdot)$ is the mixing operation as shown in Figure \ref{Fig:patchmax}. $\mathrm{y_t'}$ is the pseudo label of the given target image $\mathrm{x}_t$.  
    Note that the target pseudo label is generated from a Mean-Teacher framework, where the teacher model is the exponential moving average of the student. Since our method attaches source patches to the target image, it can effectively diversify the content of training images.

    %------------------------------ GTA5-> Cityscapes Results -------------------------------------------------

    \begin{table*}[!t]
     \centering
     \footnotesize
     \tabcolsep=2pt
     \caption{Adaptation from GTA5 to Cityscapes. $\#$ TS denotes the number of target samples used in training. The best results of one-shot domain adaptation are presented in \textbf{bold}. }\label{Tab:GTA2CS}
     \resizebox{\textwidth}{!}{
     \begin{tabular}{lccccccccccccccccccccc}
       \toprule[1.0pt]
        {Method} & $\#$TS & \rotatebox{90}{road} & \rotatebox{90}{side.} & \rotatebox{90}{build.} & \rotatebox{90}{wall}   & \rotatebox{90}{fence} & \rotatebox{90}{pole} & \rotatebox{90}{light}  & \rotatebox{90}{sign}  & \rotatebox{90}{vege.} & \rotatebox{90}{terr.}  & \rotatebox{90}{sky}   & \rotatebox{90}{person} & \rotatebox{90}{rider}  & \rotatebox{90}{car}   & \rotatebox{90}{truck} & \rotatebox{90}{bus}    & \rotatebox{90}{train} & \rotatebox{90}{motor.} & \rotatebox{90}{bike}   & \rotatebox{90}{\textbf{mIoU}} \\

      \midrule[1.0pt]

      %------------- ResNet101------------------------

       %\midrule[0.7pt]
       \multicolumn{22}{c}{\textbf{UDA}} \\
       \midrule[0.7pt]
       
       CBST \cite{CBST} & All & 91.8 & 53.5 & 80.5 & 32.7 & 21.0 & 34.0 & 28.9 & 20.4 & 83.9 & 34.2 & 80.9 & 53.1 & 24.0 & 82.7 & 30.3 & 35.9 & 16.0 & 25.9 & 42.8 & 45.9 \\
       DACS \cite{DACS} & All & 89.9 & 39.7 & 87.9 & 30.7 & 39.5 & 38.5 & 46.4 & 52.8 & 88.0 & 44.0 & 88.8 & 67.2 & 35.8 & 84.5 & 45.7 & 50.2 & 0.0 & 27.3 & 34.0 & 52.1 \\
       UPLR \cite{UncerDA} & All & 90.5 & 38.7 & 86.5 & 41.1 & 32.9 & 40.5 & 48.2 & 42.1 & 86.5 & 36.8 & 84.2 & 64.5 & 38.1 & 87.2 & 34.8 & 50.4 & 0.2 & 41.8 & 54.6 & 52.6 \\
       ProDA \cite{ProDA} & All & 87.8 & 56.0 & 79.7 & 46.3 & 44.8 & 45.6 & 53.5 & 53.5 & 88.6 & 45.2 & 82.1 & 70.7 & 39.2 & 88.8 & 45.5 & 59.4 & 1.0 & 48.9 & 56.4 & 57.5 \\
       CPSL \cite{CPSL} & All & 92.3 & 59.9 & 84.9 & 45.7 & 29.7 & 52.8 & \textbf{61.5} & 59.5 & 87.9 & 41.5 & 85.0 & 73.0 & 35.5 & 90.4 & 48.7 & 73.9 & 26.3 & 53.8 & 53.9 & 60.8 \\
       DAFormer \cite{DAFormer} & All & 95.7 & 70.2 & 89.4 & \textbf{53.5} & 48.1 & 49.6 & 55.8 & 59.4 & \textbf{89.9} & \textbf{47.9} & \textbf{92.5} & 72.2 & 44.7 & 92.3 & 74.5 & \textbf{78.2} & 65.1 & 55.9 & 61.8 & 68.3 \\
       \midrule
       \textbf{IDM (Ours)} & All & \textbf{97.2} & \textbf{77.1} & \textbf{89.8} & 51.7 & \textbf{51.7} & \textbf{54.5} & 59.7 & \textbf{64.7} & 89.2 & 45.3 & 90.5 & \textbf{74.2} & \textbf{46.6} & \textbf{92.3} & \textbf{76.9} & 59.6 & \textbf{81.2} & \textbf{57.3} & \textbf{62.4} & \textbf{69.5} \\
       
       \midrule[0.7pt]
       \multicolumn{22}{c}{\textbf{One-shot UDA}} \\
       \midrule[0.7pt]      

       CBST \cite{CBST} & One & 76.1 & 22.2 & 73.5 & 13.8 & 18.8 & 19.1 & 20.7 & 18.6 & 79.5 & 41.3 & 74.8 & 57.4 & 19.9 & 78.7 & 21.3 & 28.5 & 0.0 & 28.0 & 13.2 & 37.1 \\
       ProDA \cite{ProDA} & One & 80.9 & 32.2 & 68.9 & 24.7 & 21.0 & 24.6 & 29.6 & 14.8 & 71.7 & 28.6 & 66.4 & 55.8 & 17.5 & 81.6 & 21.2 & 24.2 & {20.0} & 25.0 & 13.9 & 38.0 \\
       ASM\cite{ASM}   & One & \textbf{89.5} & 31.2 & 81.3 & 27.8 & 22.8 & 30.6 & 32.8 & 25.1 & 82.6 & \textbf{35.0} & 76.7 & 59.2 & 26.6 & 82.3 & 27.7 & 34.1 & 0.9 & 25.6 & 29.6 & 43.2 \\
       SM-PPM \cite{SM-PPM} & One & 85.0 & 23.2 & 80.4 & 21.3 & 24.5 & 30.0 & 32.0 & 26.7 & 83.2 & 34.8 & 74.0 & 57.3 & 29.0 & 77.7 & 27.3 & {36.5} & 5.0 & 28.2 & 39.4 & 42.8 \\
       DAFormer \cite{DAFormer} & One & 88.7 & \textbf{34.4} & 84.9 & 29.1 & 28.5 & 36.9 & 43.9 & 29.7 & 83.4 & 29.6 & 84.1 & 66.0 & 38.0 & 86.8 & 54.9 & 47.3 & 32.8 & 24.6 & 37.8 & 50.6 \\
       \midrule
       
       \textbf{IDM (Ours)} & One & 88.5 & 30.0 & \textbf{86.7} & \textbf{35.0} & \textbf{33.6} & \textbf{45.0} & \textbf{49.9} & \textbf{50.7} & \textbf{86.9} & 32.8 & \textbf{86.1} & \textbf{68.1} & \textbf{40.0} & \textbf{89.1} & \textbf{66.4} & \textbf{50.6} & \textbf{45.6} & \textbf{39.3} & \textbf{52.1} & \textbf{56.7} \\ 
      \bottomrule[1.0pt]
      \end{tabular}
      }
    \end{table*}

    \noindent\textbf{Prototype-based Information Maximization.} 
    Since the original semantic structure information has been destroyed in the mixed image pair ($\mathrm{\widetilde{x}}_t, \mathrm{\widetilde{\mathrm{y}}}_t$), we adopt a metric learning method to enhance the feature representation \cite{Sepico,ProCA,ProDA}. To be specific, we use supervised contrastive learning to explore the semantic consistency between intra-class and inter-class. The proposed prototype-based supervised contrastive loss is as follows:
    \begin{equation}
        \footnotesize
        \label{Equ:ProCons}
        \mathcal{L}_{scl}(\mathrm{\widetilde{x}}_t) =- \sum_{c=1}^{C} \sum_{i=1}^{H \times W} \mathrm{\widetilde{y}}_t \log \frac{\exp (\mathrm{p}_c \cdot \mathit{F}_{\mathrm{\widetilde{x}}_t}^{(c,i)}/\tau)}{\sum_{c=1}^{C}\exp (\mathrm{p}_c \cdot \mathit{F}_{\mathrm{\widetilde{x}}_t}^{(c,i)}/\tau)},
    \end{equation}
    where $\tau$ is the temperature. $\mathit{F}_{\mathrm{\widetilde{x}}_t}^{(c,i)}$ is the representation feature of mixed image $\mathrm{\widetilde{x}}_t$ in pixel $i$ belonging to the category $c$.
    $\mathrm{p}_c$ is the prototype of category $c$, and it is calculated on the stylized image $(\mathrm{\hat{x}}_s, \mathrm{y}_s)$:
    \begin{equation}
        \label{Equ:proto}
        \footnotesize
        \mathrm{p}_c = \frac{\sum_{n=1}^{N} \sum_{i=1}^{H \times W} \mathit{F}_{\mathrm{\hat{x}}_s}^{(n,i)}\mathbbm{I}[\mathrm{y}_s^{(n, i)}=c]}{\sum_{n=1}^{N} \sum_{i=1}^{H \times W}\mathbbm{I}[\mathrm{y}_s^{(n, i)}=c]}.
    \end{equation}
    Note that our prototype is computed based on the feature $\mathit{F}_{\mathrm{\hat{x}}_s}^{(n,i)}$ of stylized images, identified by Sec. (\ref{sec:SEM}), due to the label $\mathrm{y}_s$ being the ground truth, which can remove the harmful label noise. 
    To diversify the output of the adaptation model, we also provide an information maximization loss that is formulated based on the prototypes. Details are as follows.
    \begin{equation}
        \label{Equ:IM}
        \footnotesize
        \mathcal{L}_{im}(\mathrm{\widetilde{x}}_t) = \sum_{c=1}^{C} \mathrm{\hat{p}}_c \log \mathrm{p}_{c}^{\mathrm{\widetilde{x}}_t},
    \end{equation}
    where $\mathrm{\hat{p}}_c$ is the mean prototype embedding of the whole selected source image, and $\mathrm{p}_c^{\mathrm{\widetilde{x}}_t}$ is the prototype of the mixed target image. Then we maximize the following objective for the mixed target data:
    \begin{equation}
        \label{Equ:DIM}
        \footnotesize
        \mathcal{L}_{pim}(\mathrm{\widetilde{x}}_t) = \mathcal{L}_{im}(\mathrm{\widetilde{x}}_t) - \mathcal{L}_{scl}(\mathrm{\widetilde{x}}_t).
    \end{equation}
    
    Finally, we achieve efficient one-shot domain adaptation by jointly training the sample-selected minimization and the prototype-based information maximization as follows:
    \begin{equation}
        \label{Equ:final}
        \footnotesize
        \mathcal{L}(\mathrm{\hat{x}}_s, \mathrm{\widetilde{x}}_t) = \mathcal{L}_{ssm}(\mathrm{\hat{x}}_s) + \mathcal{L}_{pim}(\mathrm{\widetilde{x}}_t).
    \end{equation}

\section{Experiments}\label{sec:exp}

\subsection{Datasets} 
\label{sec:dataset}

    \textbf{Cityscapes}, treated as target data, is a real-world dataset collected from several German cities. It has 2,975 training images with a resolution of 2048 $\times$ 1024. In our experiments, we use only one unlabeled image during training. We use the full validation set with 500 images to test our network. \textbf{GTA5}, consisting of 24,966 images, is collected from the homonymous computer game, and the original image size is 1914$\times$1052. It has 19 common categories with Cityscapes, and the ground truth is generated by the game render itself. \textbf{SYNTHIA} is another synthetic dataset that contains 9,400 fully annotated images with the original resolution of 1280 $\times$ 760. We only evaluate a subset of 13 and 16 classes common with Cityscapes.

    %----------------- SYNTHIA -> Cityscapes Results -------------------------------------

    \begin{table*}[!t]
     \centering
     \tabcolsep=2pt
     \footnotesize
     %\small
     \caption{Adaptation from SYNTHIA to Cityscapes. $\#$ TS denotes the number of target samples used in training. The best results of one-shot domain adaptation are presented in \textbf{bold}.}\label{Tab:SYN2CS}
     \resizebox{\textwidth}{!}{
     \begin{tabular}{lccccccccccccccccccc}
     \toprule[1.0pt]
     Method & $\#$TS & \rotatebox{90}{road} & \rotatebox{90}{side.} & \rotatebox{90}{build.}  &  \rotatebox{90}{wall$^{*}$}  &  \rotatebox{90}{fence$^{*}$}  &  \rotatebox{90}{pole$^{*}$} &  \rotatebox{90}{light}  & \rotatebox{90}{sign}  & \rotatebox{90}{vege.} & \rotatebox{90}{sky}   & \rotatebox{90}{person} & \rotatebox{90}{rider}  & \rotatebox{90}{car}   & \rotatebox{90}{bus}   & \rotatebox{90}{motor.} & \rotatebox{90}{bike}   & \rotatebox{90}{\textbf{mIoU}$^{*}$} & \rotatebox{90}{\textbf{mIoU}} \\

     \midrule[0.7pt]

     %------------- ResNet101------------------------

     %\midrule[0.7pt]
     \multicolumn{20}{c}{\textbf{UDA}} \\
     \midrule[0.7pt]
     CBST \cite{CBST} & All & 68.0 & 29.9 & 76.3 & 10.8 & 1.4 & 33.9 & 22.8 & 29.5 & 77.6 & 78.3 & 60.6 & 28.3 & 81.6 & 23.5 & 18.8 & 39.8 & 42.6 & 48.9 \\
     DACS \cite{DACS} & All & 80.6 & 25.1 & 81.9 & 21.5 & 2.9 & 37.2 & 22.7 & 24.0 & 83.7 & \textbf{90.8} & 67.6 & 38.3 & 82.9 & 38.9 & 28.5 & 47.6 & 48.3 & 54.8 \\
     UPLR \cite{UncerDA} & All & 79.4 & 34.6 & 83.5 & 19.3 & 2.8 & 35.3 & 32.1 & 26.9 & 78.8 & 79.6 & 66.6 & 30.3 & 86.1 & 36.6 & 19.5 & 56.9 & 48.0 & 54.6 \\
     ProDA \cite{ProDA} & All & \textbf{87.8} & 45.7 & 84.6 & 37.1 & 0.6 & 44.0 & 54.6 & 37.0 & \textbf{88.1} & 84.4 & 74.2 & 24.3 & 88.2 & 51.1 & 40.5 & 45.6 & 55.5 & 62.0 \\
     CPSL \cite{CPSL} & All & 87.2 & 43.9 & 85.5 & 33.6 & 0.3 & 47.7 & 57.4 & 37.2 & 87.8 & 88.5 & \textbf{79.0} & 32.0 & \textbf{90.6} & 49.4 & 50.8 & 59.8 & \textbf{65.3} & 57.9 \\
     DAFormer \cite{DAFormer} & All & 84.5 & 40.7 & \textbf{88.4} & \textbf{41.5} & \textbf{6.5} & 50.0 & 55.0 & 54.6 & 86.0 & 89.8 & 73.2 & 48.2 & 87.2 & 53.2 & 53.9 & 61.7 & 60.9 & 67.4 \\
     \midrule
     \textbf{IDM (Ours)} & All & 87.6 & \textbf{47.6} & 88.1 & 33.4 & 6.3 & \textbf{52.8} & \textbf{57.8} & \textbf{56.5} & 83.0 & 77.5 & 66.2 & \textbf{52.1} & 89.3 & \textbf{55.6} & \textbf{57.1} & \textbf{64.2} & 60.9 & \textbf{67.9} \\
     
     \midrule[0.7pt]
     \multicolumn{20}{c}{\textbf{One-shot UDA}} \\
     \midrule[0.7pt]
      
     CBST \cite{CBST} & One & 59.6 &  24.1 & 72.9 & - & - & - & 5.5 & 13.8 & 72.2 & 69.8 & 55.3 & 21.1 & 57.1 & 17.4 & 13.8 & 18.5 & - & 38.5 \\
     ProDA \cite{ProDA} & One & 81.8 & 38.9 & 60.6 & 7.8 & 0 & 31.6 & 14.6 & 11.5 & 51.5 & 69.9 & 56.2 & 16.4 & 79.2 & 24.4 & 5.9 & 32.3 & 36.4 & 41.8  \\
     ASM \cite{ASM} & One & \textbf{85.7} & \textbf{39.7} & 77.1 & 1.1 & 0.0 & 24.2 & 2.1 & 9.2 & 76.9 & 81.7 & 43.4 & 11.4 & 63.9 & 15.8 & 1.6 & 20.3 & 34.6 & 40.7 \\
     SM-PPM \cite{SM-PPM} & One & 79.3 & 35.3 & 75.9 & 5.6 & \textbf{16.6} & 29.8 & 25.4 & 22.7 & 79.9 & 76.8 & 54.6 & 23.5 & 60.2 & 23.9 & 21.2 & \textbf{36.6} & 41.4 & 47.3 \\
     DAFormer \cite{DAFormer} & One & 65.3 & 26.1 & 79.5 & \textbf{24.8} & 1.9 & 38.3 & 30.7 & 23.8 & 81.4 & \textbf{84.0} & \textbf{66.1} & \textbf{27.6} & 70.8 & \textbf{39.3} & 23.7 & 33.8 & 44.8 & 50.2 \\
     \midrule
     \textbf{IDM (Ours)} & One & 85.4 & 39.4 & \textbf{83.5} & 11.6 & 0.6 & \textbf{43.9} & \textbf{45.4} & \textbf{31.7} & \textbf{86.0} & 83.9 & 62.3 & 23.3 & \textbf{87.4} & 32.4 & \textbf{25.1} & 34.1 & \textbf{48.5} & \textbf{55.4} \\
     \bottomrule[1.0pt]
     \end{tabular}
     }
    \end{table*}

\subsection{Implementation Details}
\label{sec:implementation}

    We conduct all experiments by using \textbf{PyTorch} trained on the \textbf{GeForce RTX 3090Ti} GPU. 
    Following previous works \cite{DAFormer}, we adopt the transformer-based architecture as a strong baseline. For a fair comparison, we also perform all experiments on the DeepLab-v2 with ResNet101 as the backbone. We train the model with an AdamW optimizer, a learning rate of $ 6 \times 10^{-5}$ for the encoder and $6 \times 10^{-4}$ for the decoder, a weight decay of $0.01$, linear learning rate warmup with 500 iterations and linear decay afterwards. Similar to \cite{DAFormer}, rare class sampling is also applied. We first train the network with a batch of two $640 \times 640$ random crops for total 40k iterations to obtain a high-quality source model. Then it is considered the initialization model. During one-shot adaptation, we follow previous works \cite{ASM,SM-PPM} using only one target image to achieve quick adaptation. Concerning the hyper-parameters, we utilize $\lambda_{ent}=0.015$, $\lambda_{sim}=0.5$, $k=13$, and $\tau=100$ for all experiments.
    Following \cite{ASM}, we run our methods 5 times with different random seeds to get an average result, where each time we randomly select one target image for training.    
    
    %%%%%%%%%% Figure of Inference-time Adaptation %%%%%%%%
    \begin{figure*}[!t]
    \begin{minipage}[t]{0.45\linewidth} 
    \centering  
    \setlength{\belowcaptionskip}{-0.1cm}    
    \includegraphics[width=1\linewidth,trim=220 240 195 155,clip]{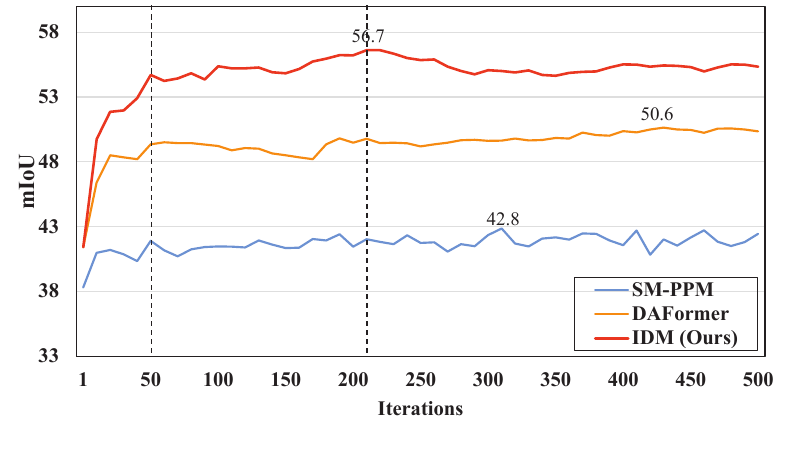}  
    \caption{The convergence during training iterations.}\label{Fig:Dis}  
    \end{minipage}%  
    \hspace{34pt}
    \begin{minipage}[t]{0.45\linewidth}  
    \centering
    \includegraphics[width=1\linewidth,trim=215 180 210 215,clip]{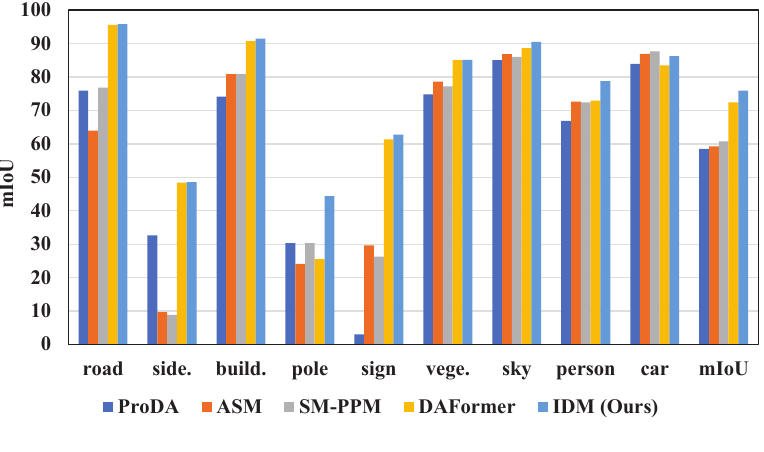} 
    \caption{Comparison results on the inference time.}  
    \label{Fig:Infer} 
    \end{minipage}  
    \end{figure*}
    %%%%%%%%%%%%%%%%%%%%%%%%%%%%%%%%%%%%%%%%%%%%%%%%%%%%%%%

\subsection{Comparison with State-of-the-art Methods}
\label{sec:sota}

    To testify the effectiveness of our method, we compare the proposed IDM with two different cross-domain semantic segmentation scenarios, including conventional unsupervised domain adaptation (UDA) and one-shot unsupervised domain adaptation (One-shot UDA). Besides, we verify the efficiency and inference-time performance of the proposed IDM.

    \noindent\textbf{Unsupervised Domain Adaptation (UDA).} We first compare our method with the conventional UDA approaches in that all target data is available during training. We compare representative state-of-the-art approaches using ResNet-101 \cite{ResNet} backbone, \textit{e.g.} CBST \cite{CBST}, DACS \cite{DACS}, UPLR \cite{UncerDA}, ProDA \cite{ProDA}, and CPSL \cite{CPSL}, and Transformer-based architecture, \textit{e.g.} DAFormer \cite{DAFormer}. The detailed results are shown in Table \ref{Tab:GTA2CS} for GTA5 to Cityscapes and Table \ref{Tab:SYN2CS} for SYNTHIA to Cityscapes. From the results, we can observe that the proposed IDM outperforms existing traditional unsupervised domain adaptation methods. Specifically, it achieves the performance of 69.5\%/67.9\% mIoU compared to DAFormer 68.3\%/67.4\% mIoU on GTA5/SYNTHIA to Cityscapes, respectively. We also provide the comparison results based on DeepLab-v2 using ResNet-101 as the backbone. More results are shown in \textit{Supplementary Materials.}

    \noindent\textbf{One-shot Unsupervised Domain Adaptation (One-shot UDA).} To further testify the potential of IDM on one-shot domain adaptation, we compare existing methods from two aspects, 1) conventional UDA methods under one-shot scenario (\textit{e.g.}, CBST \cite{CBST} and ProDA \cite{ProDA}, DAFormer \cite{DAFormer}), and 2) state-of-the-art one-shot UDA methods (\textit{e.g.}, ASM \cite{ASM}, SM-PPM \cite{SM-PPM}). From the results in Table \ref{Tab:GTA2CS} and \ref{Tab:SYN2CS}, we can draw some conclusions as follows. First, the proposed IDM outperforms the above methods under the one-shot UDA setting. Specifically, our method provides a significant improvement with 6.1\% higher than DAFormer on GTA5 $\rightarrow$ Cityscapes. Considering SYNTHIA$\rightarrow$Cityscapes task in Table \ref{Tab:SYN2CS}, our method achieves 48.5\% and 55.4\% results on 13 and 16 classes, respectively, which provides a significant margin improvement over ASM, SM-PPM and DAFormer. 
    These results demonstrate the effectiveness of the proposed informative data mining adaptation approach. Second, compared with conventional domain adaptation, the performance of one-shot UDA methods degrades significantly in such data-scarce scenarios. For example, the performance of one-shot DAFormer reduces to 50.6\% mIoU on GTA5 $\rightarrow$ Cityscapes from the original 68.3\% mIoU. It shows the naive model by directly reducing target data of UDA methods is infeasible for one-shot UDA due to the over-fitting to the single target image. 
    It also reveals the essential of our method for one-shot domain adaptation. Moreover, it should notice that we have achieved an impressive performance with only 500 training iterations, which further demonstrates our efficiency. 

\subsection{Inference-time Adaptation Performance}
\label{sec:infer-time}

    To testify the efficiency of our IDM for tackling dynamically changing environment scenarios, 
    we perform the inference-time performance comparisons in this subsection, including model convergence speed and the performance on inference-time adaptation.

\noindent\textbf{Training Convergence.} 
    We first compare the convergence speed during adaptation training, which is essential for one-shot domain adaptation to adapt fast to different real-world scenarios. As ASM \cite{ASM} requires 200k training iterations and an additional style transfer module optimization, we compare it with a more efficient method SM-PPM \cite{SM-PPM}. As Figure \ref{Fig:Dis} shows, the proposed IDM can quickly fit into the new target domain. We observe that IDM achieves a higher and more stable adaptation result by only 50 training iterations, with a significant margin gain and quick speed for convergence. Moreover, we also provide training convergence of one-shot DAFormer. Our method outperforms DAFormer both in accuracy and speed. It reveals that our IDM is superior in both efficiency and effectiveness.
    
    \noindent\textbf{Inference-time Adaptation Results.} We provide segmentation results for the referred one-shot training target data. Because the target image is unlabeled, it is reasonable to evaluate this single image during inference time, as same as test time training \cite{TENT}, which reveals the capability of our model to fit different scenarios. We report the average results on five randomly selected target images, and the compared approaches are evaluated in the same manner. We compare with ProDA \cite{ProDA}, ASM \cite{ASM}, SM-PPM \cite{SM-PPM}, and DAFormer \cite{DAFormer} in this subsection. We provide the performance of common classes among different images and the results are shown in Figure \ref{Fig:Infer}. We observe that the proposed IDM achieves the best results in most categories. Note that our method is computed based on the model trained for 50 iterations, while ProDA and ASM are trained for 200k iterations. It demonstrates that our IDM achieves quick adaptation for dealing with dynamic domain shift problems.

\begin{table}[!t]
    \centering
    \footnotesize
    \setlength{\tabcolsep}{5pt}
    \caption{Study on each component adopted by our IDM. SSM: sample-selected minimization, PIM: prototype-based information maximization.}\label{Tab:Abl}
    \begin{tabular}{lcccc|c}
    \toprule
    Network & SSM & PatchMix & ClassMix & PIM & mIoU \\
    \midrule
    {\scriptsize 1.} DAFormer \cite{DAFormer} & - & - & - & - & 50.6   \\
    {\scriptsize 2.} DAFormer \cite{DAFormer} & \checkmark & - & - & - & 52.0 \\
    {\scriptsize 3.} DAFormer \cite{DAFormer} & \checkmark & - & - & \checkmark & 53.2 \\ 
    {\scriptsize 4.} DAFormer \cite{DAFormer} & \checkmark & \checkmark & - & - &  55.0 \\
    {\scriptsize 5.} DAFormer \cite{DAFormer} & - &\checkmark & - & \checkmark & 54.8 \\
    
    {\scriptsize 6.} DAFormer \cite{DAFormer} & \checkmark & - & \checkmark &  \checkmark & 55.9 \\ %54.9
    {\scriptsize 7.} DAFormer \cite{DAFormer} & \checkmark & \checkmark & - & \checkmark & \textbf{56.7} \\
    \midrule
    {\scriptsize 8.} DLv2 \cite{DeepLab-V2} & - & - & - & - & 41.1 \\ 
    {\scriptsize 9.} DLv2 \cite{DeepLab-V2} & \checkmark & \checkmark & - & \checkmark & 45.2 \\
    \bottomrule 
    \end{tabular}
\end{table}

\subsection{Ablation Study}
\label{sec:abla}
    
    \noindent\textbf{Influence of Different Components.} In this section, we first conduct experiments to verify the effectiveness of the proposed components, including \textit{sample-selected minimization} (SSM), and \textit{prototype-based information maximization} (PIM). Specifically, we also compare the influence of the proposed \textit{PatchMix} with existing ClassMix \cite{ClassMix} in the experiments. For a fair comparison, we provide the results of two different architectures, DAFormer \cite{DAFormer} and DeepLab-v2 \cite{DeepLab-V2}. As Table \ref{Tab:Abl} shows, our model provides a significant improvement by achieving 56.7\% mIoU and 45.2 \% mIoU based on the DAFormer and DeepLab-v2, respectively, compared to the baseline model of 50.6\% and 41.1 \%. In addition, the model with SSM (model (1)) improves performance from 50.6\% to 52.0\%, which indicates the proposed sample selection technique can effectively transfer the target information into the trained model. Besides, the model with SSM and PIM can bring 1.2\% performance gain, which indicates the proposed two parts are complementary for model adaptation. In addition, adding PatchMix from SSM provides a significant improvement from 52.0\% (model (1)) to 55.0\% (model (4)). The reason is that the proposed SSM focuses on the style of information transfer and the PIM pays more attention to context information. Moreover, comparing the PatchMix and ClassMix, model (6) and model (7), our method PatchMix outperforms 0.8\% mIoU than ClassMix, owning to the rare classes' existence in the target images. Furthermore, Our method also provides a large margin improvement based on DeepLab-v2 architecture from 41.1\% to 45.2\%.

\begin{table}[!tbp]
\centering
\footnotesize
\caption{Performance on different architectures for GTA5 and SYNTHIA (SYN) to Cityscapes (CS) adaptation. Experiments are conducted on both conventional domain adaptation (UDA) and one-shot domain adaptation (OSDA).}
\label{Tab:Arch}
\setlength{\tabcolsep}{6pt}
\begin{tabular}{lcc|cc}
\toprule
 & \multicolumn{2}{c|}{GTA5 $\rightarrow$ CS} & \multicolumn{2}{c}{SYN $\rightarrow$ CS}  \\
 
 & OSDA  & UDA & OSDA & UDA      \\
\midrule
DLv2$_{\_\textit{Src. Only}}$      & 36.9  & 36.9 & 38.6 & 38.6  \\
DLv2$_{\_\textit{IDM}}$           & 45.2  & 57.3 & 47.2 & 65.9 \\
\midrule
DAFormer$_{\_\textit{Src. Only}}$  & 44.5 & 44.5 & 51.3 & 51.3  \\
DAFormer$_{\_\textit{IDM}}$       & 56.7 & 69.5 & 55.4 & 67.9  \\
\bottomrule
\end{tabular}
\end{table}

    \noindent\textbf{Performance on Different Architectures.} To verify the generalization of our method, we perform the experiment on different architectures, including the traditional convolution-based architecture DeepLab-v2 \cite{DeepLab-V2} and advanced transformer-based architecture DAFormer \cite{DAFormer}. Specifically, we provide the conventional domain adaptation (UDA) and one-shot domain adaptation (OSDA) results on Table \ref{Tab:Arch}. From the results, we can observe that the proposed method (IDM) offers a significant improvement on both GTA5 to Cityscapes (GTA5$\rightarrow$CS) and SYNTHIA to Cityscapes (SYN$\rightarrow$CS) adaptations. The detailed results are attached in the \textit{Supplementary Materials}.   

\noindent\textbf{Influence of Different Selection Strategies.}
    To identify the most informative samples, we have proposed two different selection strategies: prediction uncertainty selection ($\mathcal{W}^{p}$) and similarity uncertainty selection ($\mathcal{W}^{s}$). As our sample section is based on the style transferred (ST) images, we ablate both of them  in this subsection. We conduct experiments on GTA5 $\rightarrow$ Cityscapes to verify the effectiveness of different strategies. The detailed results are shown in Table \ref{Tab:Sample}. 
    Compared with the baseline, introducing style transfer brings performance improvement from 50.6\% to 51.5\%, without reducing significant training iterations. Besides, adding prediction and similarity uncertainty selection techniques, the model achieves 52.0\% mIoU with only 300 training iterations. This reveals our method is efficient to perform quick adaptation and verifies the effectiveness of the proposed informative sample selection strategy.

\begin{table}[!t]
    \centering
    \footnotesize
    \setlength{\tabcolsep}{8pt}
    \caption{Performance on different sample selection strategies for GTA5 to Cityscapes adaptation.}
    \label{Tab:Sample}
    \begin{tabular}{cccccc}
    \toprule
     Method & ST & $\mathcal{W}^p$ & $\mathcal{W}^s$ & mIoU & Iterations \\
    \midrule
    Baseline  & - & - & - & 50.6 & 20000 \\
    \midrule
    (a) & \checkmark & - & - & 51.5 & 13000 \\
    (b) & - & \checkmark & -  & 51.4 & 2000 \\
    (c) & \checkmark & - & \checkmark & 51.7 & 4300 \\
    (d) & \checkmark & \checkmark & \checkmark & 52.0 & 300 \\
    \bottomrule
    \end{tabular}
\end{table}    

\begin{table}[!tbp]

\begin{minipage}{0.3\linewidth}
\centering
\caption{
Study on the uncertainty threshold parameter $\lambda_{ent}$.
}
\label{tab:lambda_ent}
\setlength{\tabcolsep}{6pt}
\begin{tabular}{cc} 
\toprule
$\lambda_{ent}$ & mIoU \\
\midrule
0.005 & 50.8 \\
0.010 & 51.1 \\
0.015 & 51.4 \\
0.020 & 50.9 \\
0.025 & 51.2 \\
\bottomrule
\end{tabular} 
\end{minipage}
\hspace{4pt}
\begin{minipage}{0.3\linewidth}
\centering
\caption{
Study on the similarity threshold parameter $\lambda_{sim}$.
}
\label{tab:lambda_sim}
\setlength{\tabcolsep}{6pt}
\begin{tabular}{cc} 
\toprule
$\lambda_{sim}$ & mIoU \\
\midrule
0.8 & 51.0 \\
0.7 & 51.6 \\
0.6 & 51.7 \\
0.5 & 52.0 \\
0.4 & 51.4 \\
\bottomrule
\end{tabular} 
\end{minipage}
\hspace{4pt}
\begin{minipage}{0.3\linewidth}
\centering
\caption{
Study on the number $k$ of categories contained in the image.
}
\label{tab:k}
\setlength{\tabcolsep}{6pt}
\begin{tabular}{cc} 
\toprule
$k$ & mIoU \\
\midrule
10 & 50.5 \\
11 & 50.7 \\
12 & 51.5 \\
13 & 51.6 \\
14 & 51.6 \\
\bottomrule
\end{tabular} 
\end{minipage}
\end{table}

    \subsection{Parameters Analysis} 
    Our framework contains several new hyper-parameters, including the prediction uncertainty selection threshold $\lambda_{ent}$ in Eq. (\ref{Eq:entropy_es}), the similarity uncertainty selection threshold $\lambda_{sim}$ in Eq. (\ref{Eq:diver_es}), and the number $k$ of categories contained in the selected stylized images in Eq. (\ref{Eq:diver_es}). We construct extensive experiments to analyze the influence of these hyper-parameters. The detailed results are provided in Table \ref{tab:lambda_ent}, \ref{tab:lambda_sim}, and \ref{tab:k}. From the results, we can observe that although our method requires many manually-defined thresholds, the performance of IDM is stable and not sensitive to these hyper-parameters. 

    Moreover, we also provide the analysis of the number of patches $P$ in the PatchMix. The key idea of PatchMix is to increase the diversity of training samples, so we randomly mix patches without specifying the corresponding positions replacement. This approach indeed somewhat breaks the semantic relations while remaining locally structured information. Fortunately, due to supervised information for training, the model can extract generalized features regardless of their spatial locations. This random mixing strategy promotes robustness and generalization in the model adaptation. The detailed results of patches $P$ are shown in the Table \ref{Tab:P}. 
\begin{table}[!tbp]
    \centering
    \caption{Study on the number of patches in the PatchMix module.}
    \label{Tab:P}
    \setlength{\tabcolsep}{7.5pt}
    \begin{tabular}{lcccccc}
    \toprule
    \# $P$ & 16 & 36 & 48 & 64 & 96 & 144  \\
    IDM & 55.8 & 56.0 & 56.1 & 56.5 & 56.7 & 56.3 \\
    \bottomrule
    \end{tabular}
\end{table}

\section{Conclusion}
\label{sec:conclusion}

     This paper proposes an Informative Data Mining (\textbf{IDM}) framework, aiming at performing quick adaptation from the pre-trained source model to the target domain by only hundreds of training iterations with one-shot target data available. To achieve this goal, we first propose a novel sample selection criterion to identify the most informative samples for training reducing redundant training significantly. 
     At the same time, we update the adaptation model by the proposed model adaptation method. Specifically, we use the prototype-based information maximization loss to enlarge the diversity of the training samples alleviating the over-fitting problem. The sample-selected minimization loss enforces the pre-trained source model to fit the target data.
     The efficacy and efficiency of IDM have been demonstrated by achieving a new state-of-the-art performance on two standard one-shot domain adaptive semantic segmentation benchmarks. 

\section{Acknowledgement}

    \noindent \quad \quad This work was supported in part by the Major Project for New Generation of AI (No. 2018AAA0100400), the National Natural Science Foundation of China (No. 61836014, No. U21B2042, No. 62072457, No. 62006231), and the InnoHK program.

{\small

\bibliographystyle{ieee_fullname}
\bibliography{IDM}
}
\end{document}